\documentclass[10pt,twocolumn,letterpaper]{article}

\usepackage{cvpr}
\usepackage{times}
\usepackage{epsfig}
\usepackage{graphicx}
\usepackage{amsmath}
\usepackage{amssymb}
\usepackage{multirow}
\usepackage{pifont}
\usepackage{booktabs}
\usepackage{epsfig}
\usepackage{booktabs,multirow}
\usepackage{graphics}
\usepackage{threeparttable}
\usepackage{color}
\usepackage[normalem]{ulem}
\usepackage{multirow}
\usepackage{float}
\usepackage{amsfonts}
\usepackage{bm}
\usepackage{subfig}
\usepackage{enumitem}
\usepackage[numbers]{natbib}
\usepackage{array}
\usepackage[table]{xcolor}
\usepackage{colortbl}
\definecolor{bblue}{rgb}{0,150,230}
\definecolor{mygray}{gray}{.9}
\definecolor{myy}{RGB}{126,95,0}
\usepackage{bm}

\newcolumntype{I}{!{\vrule width 1pt}}

\definecolor{ggray}{RGB}{127,127,127}

\makeatletter
\newcommand{\thickhline}{%
    \noalign {\ifnum 0=`}\fi \hrule height 1pt
    \futurelet \reserved@a \@xhline
}
\makeatother
\newcommand{\tabincell}[2]{\begin{tabular}{@{}#1@{}}#2\end{tabular}}
\newcommand{\figref}[1]{Fig.~\ref{#1}}
\newcommand{\tabref}[1]{Table~\ref{#1}}

\usepackage{caption}
\usepackage[pagebackref=true,breaklinks=true,letterpaper=true,colorlinks,bookmarks=false]{hyperref}

\usepackage[utf8]{inputenc}

\usepackage{cleveref}
\crefname{section}{§}{§§}
\Crefname{section}{§}{§§}
\cvprfinalcopy 

\def\cvprPaperID{4277} 

\ifcvprfinal\fi
\begin{document}

\title{Cascaded Human-Object Interaction Recognition}

\author{Tianfei Zhou$^{1}$\thanks{The first two authors contribute equally to this work.}~, Wenguan Wang$^{2*}$, Siyuan Qi$^3$, Haibin Ling$^4$, Jianbing Shen$^{1}$\thanks{Corresponding author: \textit{Jianbing Shen}.}\\
\small{$^1$Inception Institute of Artificial Intelligence, UAE~~~$^2$ETH Zurich, Switzerland~~~$^3$Google, USA~~~$^4$Stony Brook University, USA}\\
\small{\texttt{\{ztfei.debug,wenguanwang.ai\}@gmail.com}
} \\
\small\url{https://github.com/tfzhou/C-HOI}
}

\maketitle

\begin{abstract}
Rapid progress has been witnessed for human-object interaction (HOI) recognition, but most existing models are confined to single-stage reasoning pipelines. Considering the intrinsic complexity of the task, we introduce a cascade architecture for a multi-stage, coarse-to-fine HOI understanding. At each stage, an instance localization network progressively refines HOI proposals and feeds them into an interaction recognition network. Each of the two networks is also connected to its predecessor at the previous stage, enabling cross-stage information propagation. The interaction recognition network has two crucial parts: a relation ranking module for high-quality HOI proposal selection and a triple-stream classifier for relation prediction. With our carefully-designed human-centric relation features, these two modules work collaboratively towards effective interaction understanding. Further beyond relation detection on a bounding-box level, we make our framework flexible to perform fine-grained pixel-wise relation segmentation; this provides a new glimpse into better relation modeling.
%
Our approach reached the $1^{st}$ place in the ICCV2019 Person in Context Challenge, on both relation detection and segmentation tasks. It also shows promising results on V-COCO.
\end{abstract}

\vspace{-3pt}
\section{Introduction}
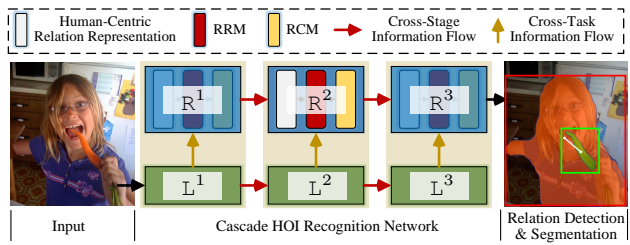
\begin{figure}[t]
		\begin{center}
			\includegraphics[width=\linewidth]{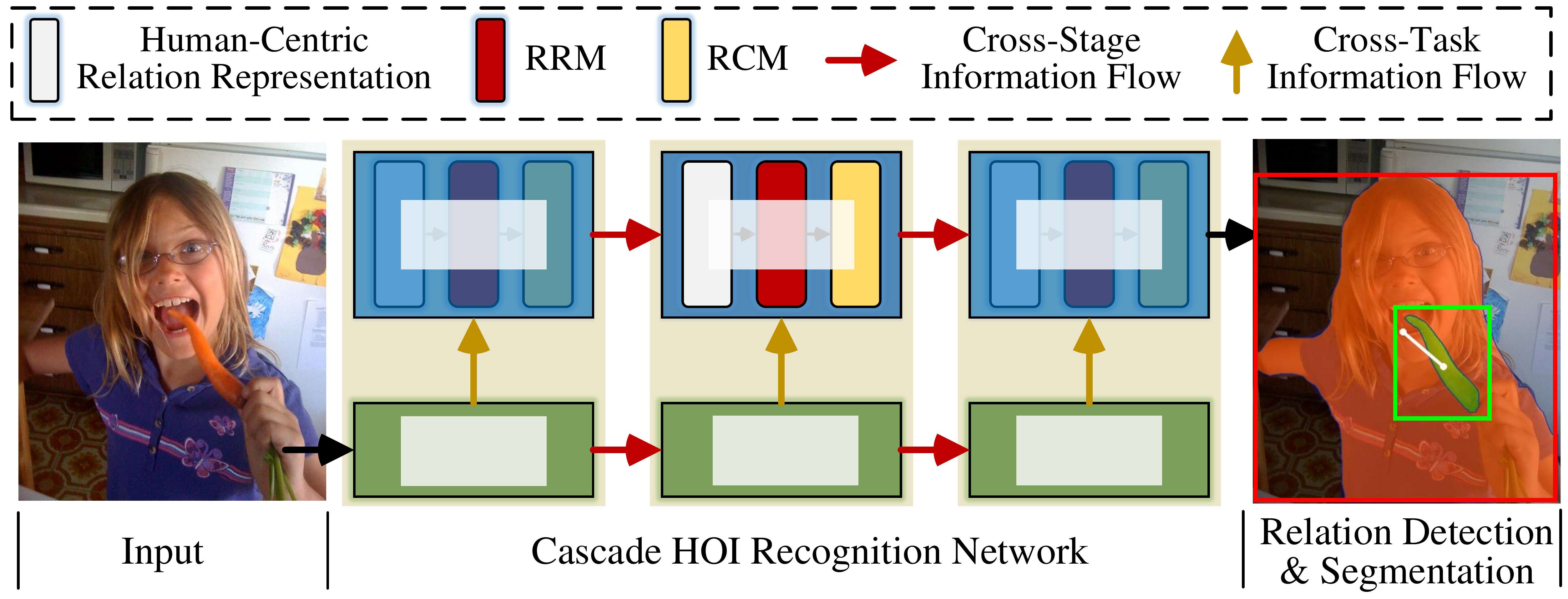}
        \put(-170,51){\small $\texttt{R}^1$}
        \put(-170,19){\small $\texttt{L}^1$}
        \put(-122,51){\small $\texttt{R}^2$}
        \put(-122,19){\small $\texttt{L}^2$}
        \put(-76,51){\small $\texttt{R}^3$}
        \put(-76,19){\small $\texttt{L}^3$}
		\end{center}
		\vspace{-16pt}
		\captionsetup{font=small}
		\caption{\textbf{Illustration of our cascade HOI recognition network,} which is able to handle both object-level relation detection and pixel-wise relation segmentation tasks. Given an input image, our model performs coarse-to-fine inference over both instance localization ($\texttt{L}^1\!\sim\!\texttt{L}^3$) and interaction recognition ($\texttt{R}^1\!\sim\!\texttt{R}^3$).}
		\vspace{-13pt}
		\label{fig:motivation}
	\end{figure}

\vspace{-2pt}
Human-object interaction (HOI) recognition aims to identify$_{\!}$ meaningful$_{\!}$ $_{\!}\left\langle\textit{{human}, {verb}, {object}}\right\rangle_{\!}$ triplets$_{\!}$ from images, such as $_{\!}\left\langle\textit{{human}, {eat}, {carrot}}\right\rangle_{\!}$ in \figref{fig:motivation}. It plays a crucial role in many vision tasks, \eg, visual question answering~\!\cite{norcliffe2018learning,li2019relation,zheng2019reasoning}, human-centric understanding~\!\cite{wang2018attentive,Wang_2019_ICCV,zhoucvpr2020}, image generation~\!\cite{johnson2018image}, and  activity recognition~\!\cite{shao2018find,wu2019learning,fan2019understanding,pang2018deep,ma2018attend}, to name a few representative ones.


Though great advances have been made recently, the task is still far from being solved. One of the main challenges comes from
its intrinsic complexity: a successful HOI recognition model must accurately 1) localize and recognize each interacted entity (\textit{human}, \textit{object}), and 2) predict the interaction classes (\textit{verb}). Both subtasks are difficult, leading to HOI recognition itself a highly complex problem.
With a broader view of other computer vision and machine learning related fields, coarse-to-fine and cascade inference have been shown to deal well with complex problems~\cite{kirkpatrick1983optimization,felzenszwalb2010cascade,felzenszwalb2006efficient,wang2019iterative}. The central idea is to leverage sequences of increasingly fine approximations to control the complexity of learning and inference. This motivates us to propose a cascade HOI recognition model, which builds up multiple stages of neural network inference in an annealing-style. For the two subtasks of instance localization and interaction recognition, this model arranges them in a successive manner within each single stage, and carries out cascade, cross-stage inference for each. Above designs result in a multi-task, coarse-to-fine inference framework, which enables asymptotically improved HOI representation learning. This also distinctively differentiates our method from previous efforts, which rely on single-stage architectures.

As shown in Fig.~\!\ref{fig:motivation}, our model consists of an instance localization network and an interaction recognition network, both working in a cascade manner. Through the instance localization network, the model step-by-step increases the selectiveness of the instance proposals. With such progressively refined HOI candidates, as well as the useful relation representation from the preceding stage, better action predictions can be achieved by current-stage interaction recognition network. Moreover, in the interaction recognition network, both human semantics and facial patterns are mined to boost relation reasoning, as these cues are 
tied to underlying purposes of human actions. With such human-centric features, a relation ranking module (RRM) is proposed to rank all the possible \textit{human}-\textit{object} pairs. Only the top-ranked, high-quality candidates are fed into a relation classification module (RCM) for final \textit{verb} prediction.

More essentially, previous HOI literature mainly address \textit{relation detection}, \ie, recognizing HOIs at a bounding-box level. In addition to addressing this classic setting, we take a further step towards more fine-grained HOI understanding, \ie, identifying the relations between interacted entities at the pixel level (see Fig.~\ref{fig:motivation}).
Studying such \textit{relation segmentation} setting not only further demonstrates the efficacy and flexibility of our cascade framework, but allows us to explore more powerful relation representations. This is because bounding box based representations only encode coarse object information with noisy backgrounds, while pixel-wise mask based features may capture more detailed and precise cues. We empirically study the effectiveness of bounding box and pixel-wise mask based relation representations as well as their hybrids. Our results suggest that the pixel-mask representation is indeed more powerful.

		
Our model reached the ${1}^{st}$ place in \textbf{ICCV-2019 Person in Context Challenge}\footnote{\scriptsize\url{http://picdataset.com/challenge/leaderboard/pic2019}} (PIC$_{19}$ Challenge), on both \textit{Human-Object Interaction in the Wild (HOIW)} and \textit{Person in Context (PIC)} tracks, where HOIW addresses relation detection, while PIC focuses on relation segmentation. Besides, it also obtains promising results on V-COCO~\cite{gupta2015visual}.

This paper makes three major contributions.
\textbf{First}, we formulate HOI recognition as a coarse-to-fine inference procedure with a novel cascade architecture.
\textbf{Second}, we introduce several techniques to learn rich features that represent the semantics of HOIs. \textbf{Third}, for the first time, we study the feature representations of HOI and find pixel-mask to be more powerful than the traditional bounding-box representation. We expect such a study could inspire more future efforts towards pixel-level HOI understanding.

\vspace{-1pt}
\section{Related Work}
\vspace{-1pt}
\noindent\textbf{Human-object interaction recognition} has a rich study history in computer vision. Early methods~\cite{yao2010modeling,yao2010grouplet,delaitre2011learning} mainly exploited human-object contextual information in structured models, such as Bayesian inference~\cite{gupta2007objects,gupta2009observing}, and compositional framework~\cite{desai2012detecting}.

\begin{figure}[t]
		\begin{center}
			\includegraphics[width=0.9\linewidth]{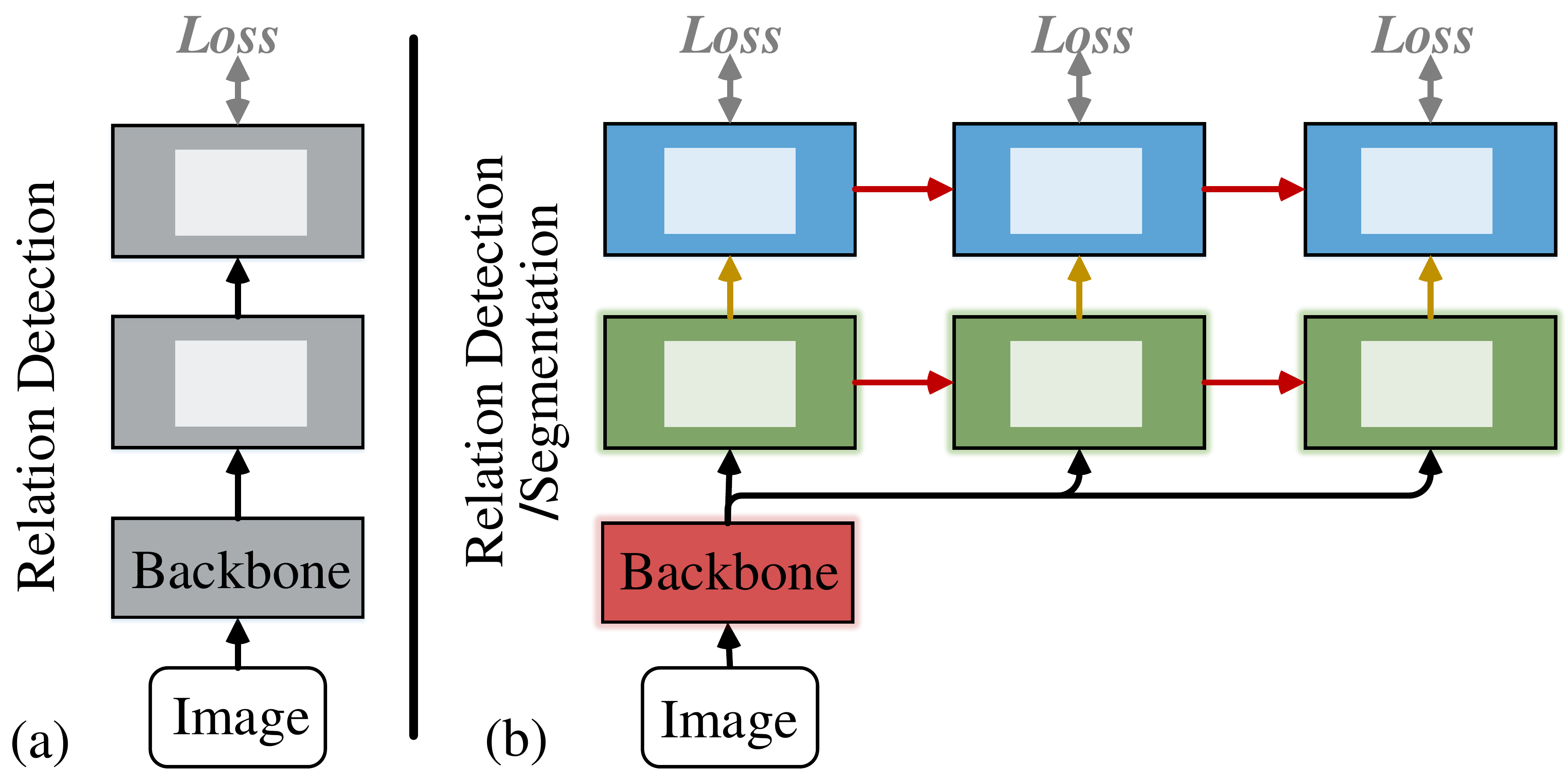}
		\put(-183,77){\small $\texttt{R}$}
        \put(-183,51){\small $\texttt{L}$}
        \put(-118,77){\small $\texttt{R}^1$}
        \put(-118,50){\small $\texttt{L}^1$}
        \put(-71,77){\small $\texttt{R}^2$}
        \put(-71,50){\small $\texttt{L}^2$}
        \put(-23,77){\small $\texttt{R}^3$}
        \put(-23,50){\small $\texttt{L}^3$}
		\end{center}
		\vspace{-10pt}
		\captionsetup{font=small}
		\caption{\small (a) Previous HOI recognition models largely rely on a single-stage architecture and only concern relation detection. (b) Our proposed HOI recognition model carries out instance localization and interaction recognition in a unified cascade architecture, and addresses both relation detection and segmentation. Note that the loss for the instance localization part is omitted for clarity. }
		\vspace{-10pt}
		\label{fig:cascade}
	\end{figure}

With the recent renaissance of neural networks in computer vision, deep learning based solutions are now dominant in this field. For instance, in~\cite{gkioxari2018detecting}, a multi-branch architecture was explored to address human, object, and relation representation learning. Some researchers revisited the classic graph model and solved this task in a neural message passing framework~\cite{qi2018learning}. For learning more effective human feature representations, pose cues have been widely adopted in recent leading approaches~\cite{li2019transferable,gupta2019no,wan2019pose,fang2018pairwise,Zhou_2019_ICCV}.
Some other efforts addressed long-tail distribution and zero-shot problems with external knowledge~\cite{gu2019scene,kato2018compositional,zhuang2018hcvrd,shen2018scaling}.
All these models use single-stage pipelines for inference (Fig.~\!\ref{fig:cascade}~\!(a)), and they can potentially benefit from the general architecture we propose here: a multi-stage pipeline that performs coarse-to-fine inference as shown in Fig.~\!\ref{fig:cascade}~\!(b).

\noindent\textbf{Object detection} has gained remarkable progress recently, benefiting from the availability of large-scale datasets (\eg, MS-COCO~\cite{lin2014microsoft}) and strong representation power of deep neural networks.
Mainstream methods are often categorized into two-stage~\cite{ren2015faster,he2017mask,chen2019hybrid,cai2019cascade} or single-stage~\cite{redmon2016you,liu2016ssd,li2017fully,law2018cornernet} paradigms. Recently, some multi-stage pipelines have been explored for coarse-to-fine object detection~\cite{cai2018cascade,chen2019hybrid}. Similarly, we revisit the general idea of cascade inference in HOI recognition, where both instance localization and relation recognition are coupled for step-by-step HOI reasoning.

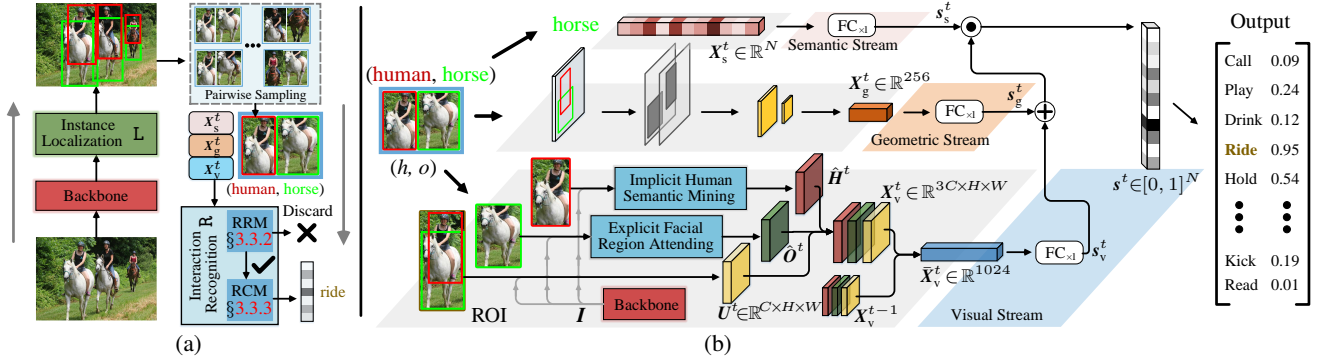
\begin{figure*}[t]
	\begin{center}
		\includegraphics[width=\linewidth]{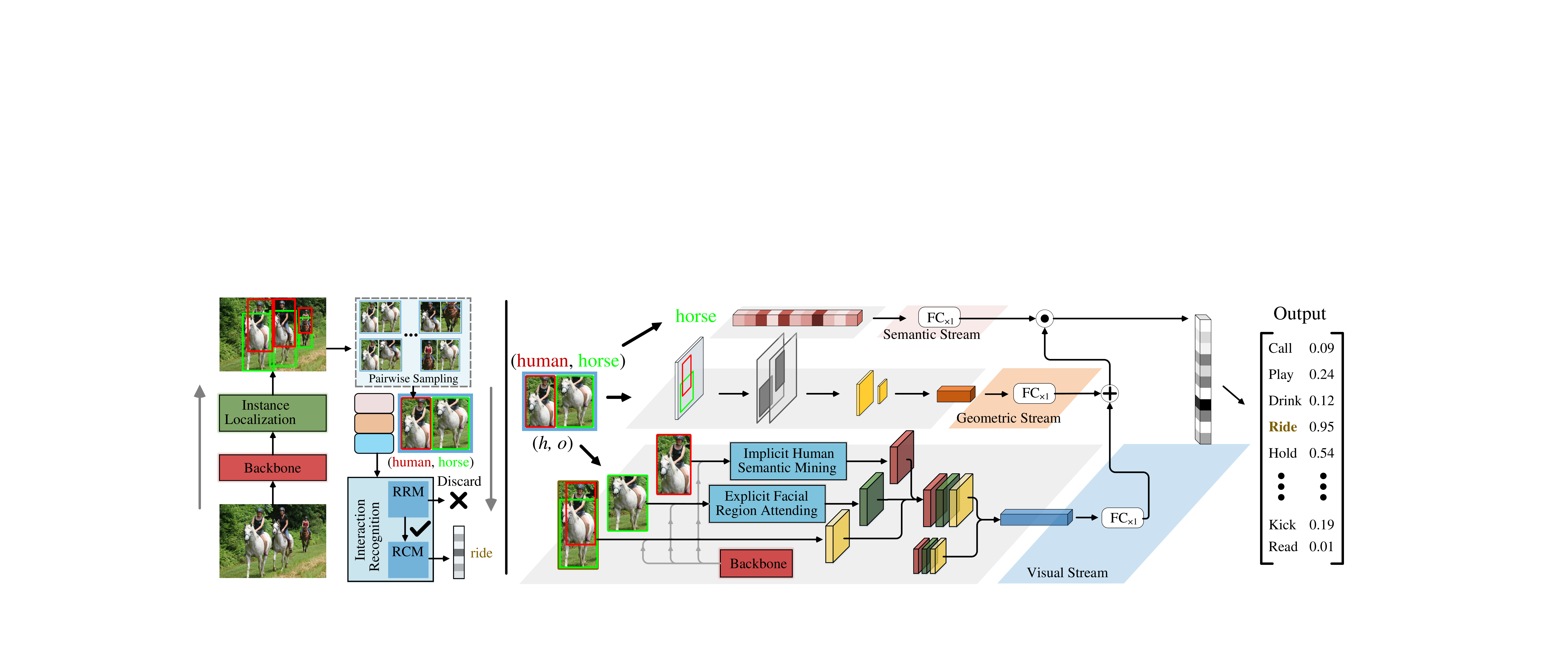}
		\put(-230,103){\scriptsize$\textbf{\textit{X}}^t_{\text{s}}\!\in\!\mathbb{R}^N$}
		\put(-177,91){\scriptsize$\textbf{\textit{X}}^t_{\text{g}}\!\in\!\mathbb{R}^{256}$}
		\put(-165,50){\scriptsize$\textbf{\textit{X}}^t_{\text{v}}\!\in\!\mathbb{R}^{3C\!\times\!H\!\times\!W}$}
		\put(-175,3){\scriptsize$\textbf{\textit{X}}^{t-1}_{\text{v}}$}
		\put(-150,18){\scriptsize$\bar{\textbf{\textit{X}}}^t_{\text{v}}\!\in\!\mathbb{R}^{1024}$}
		\put(-413,32){\scriptsize\S\ref{rrn}}
		\put(-413,6){\scriptsize\S\ref{icn}}
		\put(-185,55){\scriptsize$\hat{\textbf{\textit{H}}}^t$}
		\put(-202,25.5){\scriptsize$\hat{\textbf{\textit{O}}}^t$}
		\put(-227,4){\scriptsize$\textbf{\textit{U}}^{t\!}\!\!\in\!\!\mathbb{R}^{\!C\!\times\!H\!\times\!W}$}
		\put(-85,28){\scriptsize $\textbf{\textit{s}}^t_\text{v}$}
		\put(-117,87){\scriptsize $\textbf{\textit{s}}^t_\text{g}$}
		\put(-144,118){\scriptsize $\textbf{\textit{s}}^t_\text{s}$}
		\put(-77,53){\scriptsize $\textbf{\textit{s}}^t\!\!\in\!\![0,1]^{N}$}
		\put(-448,72){\footnotesize$\mathtt{L}$}
		\put(-423,39){\footnotesize\rotatebox{90}{$\mathtt{R}$}}
		\put(-280,3){\footnotesize{$\textbf{\textit{I}}$}}
        \put(-320,3){\footnotesize{ROI}}
		\put(-421,77){\tiny $\textbf{\textit{X}}^t_{\text{s}}$}
		\put(-421,68){\tiny $\textbf{\textit{X}}^t_{\text{g}}$}
		\put(-421,59){\tiny $\textbf{\textit{X}}^t_{\text{v}}$}
        \put(-432,-8){\small (a)}
	    \put(-232,-8){\small (b)}
	\end{center}
	\vspace{-16pt}
	\captionsetup{font=small}
	\caption{\small (a) Pipeline of our cascade network for identifying a triplet of
	$_{\!}\left\langle\textit{{\color{red}{human}}, {\color{myy}{verb}}, {\color{green}{object}}}\right\rangle_{\!}$ from an input image. (b) Illustration of our triple-stream relation classification module (RCM) that achieves HOI recognition based on our human-centric relation representation.
}
	\vspace{-8pt}
	\label{fig:framework}
\end{figure*}

\vspace{-3pt}
\section{Our Algorithm}
\vspace{-1pt}

\subsection{Cascade Network Architecture}\label{cascade}
\vspace{-2pt}
To identify $_{\!}\left\langle\textit{{human}, {verb}, {object}}\right\rangle_{\!}$ triplets in images, our method carries out progressive refinement on instance localization and relation recognition at multiple stages (see Fig.~\!\ref{fig:cascade}~\!(b)). At each stage $t$, the multi-tasking is achieved by two networks: an instance localization network $\mathtt{L}^{t\!}$ generates \textit{human} and \textit{object} proposals, and an instance recognition network $\mathtt{R}^{t\!}$ identifies the action (\ie, \textit{verb}) for each human-object pair sampled from the proposals, as shown in Fig.~\!\ref{fig:framework}~\!(a). Our cascade network is organized as follows:
\vspace{-3pt}
\begin{equation*}\small
\begin{aligned}
	\text{Instance Localization (\S\ref{oln}):}     &~~\mathcal{O}^t = \mathtt{L}^t(\mathcal{O}^{t-1}), \\
	\text{Human-Object Pair Sampling:}   &~~(h,o)\sim\mathcal{O}^t\times\mathcal{O}^t,\\
	\text{Interaction Recognition (\S\ref{irn}):} &~~\textbf{\textit{s}}^t = \mathtt{R}^t(\textbf{\textit{X}}^{t}, \textbf{\textit{X}}^{t-1}).
\end{aligned}
\vspace{-2pt}
\end{equation*}
At stage $t$, $\mathtt{L}^t$ takes the detection results $\mathcal{O}^{t-1\!}$ from $\mathtt{L}^{t-1\!}$ as inputs and outputs refined results $\mathcal{O}^{t\!}$. Then, a human-object pair $(h,o)$ is sampled from $\mathcal{O}^{t}\!\times\!\mathcal{O}^{t}$. 
Finally, $\mathtt{R}^{t}$ uses the relation features $\textbf{\textit{X}}^{t\!}$ and $\textbf{\textit{X}}^{t-1\!}$ of $(h,o)$ at current and previous stages to estimate a \textit{verb} score vector $\textbf{\textit{s}}^t$. More details about the relation feature are given in \S\ref{visual}.
Notably, 
the instance localization $\mathtt{L}^{t\!}$ and interaction recognition $\mathtt{R}^{t\!}$ networks work closely at each stage, and
$\mathtt{R}^{t\!}$ can benefit from the improved localization results $\mathcal{O}^t$ of $\mathtt{L}^t$ and give better interaction predictions.


Next we will describe in detail our instance localization network in \S\ref{oln} and interaction recognition network in \S\ref{irn}. 

\vspace{-0pt}
\subsection{Instance Localization Network} \label{oln}
\vspace{-1pt}
The instance localization network $\mathtt{L}$ outputs a set of human and object regions, from which human-object pair candidates are sampled and fed into the interaction recognition network $\mathtt{R}$ for relation classification. It is built on a cascade of detectors, \ie, at stage $t$, $\mathtt{L}^{t}$ refines an object region $o^{t-1}\!\in\!\mathcal{O}^{t-1}$ detected from the preceding stage by:
\vspace{-3pt}
\begin{equation}\small
\begin{aligned}
	~~~~~\textbf{\textit{Y}}^t = \mathtt{P}(\textbf{\textit{I}}, o^{t-1}),
\end{aligned}
	\label{eq:y}
	\vspace{-6pt}
\end{equation}
\begin{equation}
\begin{aligned}
	o^t = \mathtt{D}^t(\textbf{\textit{Y}}^t),
\end{aligned}
	\label{eq:o}
	\vspace{-1pt}
\end{equation}
where $\textbf{\textit{I}}$ is the CNN feature of the backbone network, shared by different stages. $\textbf{\textit{Y}}^t\!\in\!\mathbb{R}^{C\!\times\!H\!\times\!W}$ indicates the box feature derived from $\textbf{\textit{I}}$ and the input RoI. $\mathtt{P}$ and $\mathtt{D}^t$ represent RoIAlign~\cite{he2017mask} and a box regression head, respectively.

Similar to previous cascade object detectors~\cite{cai2018cascade,chen2019hybrid}, at each stage, $\mathtt{L}^t$ is trained with a certain interaction over union (IoU) threshold, and its output is re-sampled to train the next detector $\mathtt{L}^{t+1}$ with a higher IoU threshold.
In this way, we gradually increase the quality of training data for deeper stages in the cascade, thus boosting the selectiveness against hard negative examples.  At each stage, the instance localization loss $\mathcal{L}^t_{\text{LOC}}$ is the same as Faster R-CNN~\cite{ren2015faster}.

\vspace{-4pt}
\subsection{Interaction Recognition Network} \label{irn}
\vspace{-1pt}
As shown in \figref{fig:framework}~\!(a), the interaction recognition network $\mathtt{R}$ comprises a relation ranking module (RRM, \S\ref{rrn}) and a relation classification module (RCM, \S\ref{icn}). Both RRM and RCM rely on our elaborately designed human-centric relationship representation (\S\ref{visual}).
\vspace{-6pt}
\subsubsection{Human-Centric Relation Representation}\label{visual}
\vspace{-3pt}
At each stage $t$, for each human-object pair $(h^t,o^t)\!\in\!\mathcal{O}^t\!\times\!\mathcal{O}^t$, three types of features, \ie, \textit{semantic} feature $\textbf{\textit{X}}^t_{\text{s}}$, \textit{geometric} feature $\textbf{\textit{X}}^t_{\text{g}}$ and \textit{visual} feature $\textbf{\textit{X}}^t_{\text{v}}$,  are considered for a thorough relation representation, as shown in \figref{fig:framework}~\!(b). In the following paragraphs, the superscript `\textit{t}' is omitted for conciseness unless necessary.

\noindent\textbf{Semantic feature} $\textbf{\textit{X}}_{\text{s}}$. It captures our prior knowledge of \textit{object affordances}~\!\cite{gibson2014ecological} (\eg, a phone affords calling). We build $\textbf{\textit{X}}_{\text{s}\!}\!\in\!\mathbb{R}^{N\!}$ as the frequency of label co-occurrence between object and action categories~\!\cite{zellers2018neural}, where $N$ denotes the number of pre-defined actions in a HOI dataset.

\noindent\textbf{Geometric feature} $\textbf{\textit{X}}_{\text{g}}$. It characterizes the spatial relationship between human and object. Similar to~\!\cite{chao2018learning,gao2018ican}, we first adopt a two-channel mask representation strategy, obtaining a $(2,64,64)$-$d$ feature tensor for the two entities. Then two $\text{conv+pooling}$ operations followed by a fully connected (FC) layer are applied on the tensor to get $\textbf{\textit{X}}_{\text{g}}\!\in\!\mathbb{R}^{256}$.

\noindent\textbf{Visual feature} $\textbf{\textit{X}}_{\text{v}}$. Compared with $\textbf{\textit{X}}_{\text{s}}$ and $\textbf{\textit{X}}_{\text{g}}$, the visual feature is of greater significance and has profound effects for human beings to recognize subtle interactions. For each human-object pair $(h,o)$, we have three features $\textbf{\textit{H}}\!\in\!\mathbb{R}^{C\!\times\!H\!\times\!W}$, $\textbf{\textit{O}}\!\in\!\mathbb{R}^{C\!\times\!H\!\times\!W}$ and $\textbf{\textit{U}}\!\in\!\mathbb{R}^{C\!\times\!H\!\times\!W}$ from the human, object and their union regions correspondingly:
\vspace{-2pt}
\begin{equation}\small
	\textbf{\textit{H}} = \mathtt{P}(\textbf{\textit{I}}, h),~~~~\textbf{\textit{O}} = \mathtt{P}(\textbf{\textit{I}}, o),~~~~\textbf{\textit{U}} = \mathtt{P}(\textbf{\textit{I}}, (h,o)).
	\vspace{-1pt}
\end{equation}
Here $\textbf{\textit{H}}$, $\textbf{\textit{O}}$ and $\textbf{\textit{U}}$ are specific instances of the RoIAlign feature $\textbf{\textit{Y}}$ in Eqs.~\!(\ref{eq:y},~\!\ref{eq:o}), which are renamed to make it clear that they come from different regions.

To better capture the underlying semantics in HOI, we introduce two feature-enhancement mechanisms: \textit{implicit human semantic mining} to improve the human feature $\textbf{\textit{H}}$ and \textit{explicit facial region attending} to enhance the object feature $\textbf{\textit{O}}$. Then we have the visual feature as:
\vspace{-3pt}
\begin{equation}\small
\textbf{\textit{X}}_{\text{v}} = [\bar{\textbf{\textit{H}}},~\bar{\textbf{\textit{O}}},~{\textbf{\textit{U}}}] \in\mathbb{R}^{3C\!\times\!H\!\times\!W},
\vspace{-1pt}
\end{equation}
where $\bar{\textbf{\textit{H}}}$ and $\bar{\textbf{\textit{O}}}$ denote the enhanced human and object features, respectively, and $[\cdot]$ is the concatenation operation. Next we detail our two feature-enhancement mechanisms.

%

\noindent\textbf{1)} \textit{Implicit Human Semantic Mining.}
To reason about human-object interactions, it is essential to understand \textbf{how} humans interact with the world, \ie, which human parts are involved for an action.
Different from current leading methods resorting to expensive human pose annotations~\cite{wan2019pose,fang2018pairwise,li2019transferable}, we propose to implicitly learn human parts and their mutual interactions.

For each pixel (position) $i$ inside the human region (feature) $\textbf{\textit{H}}$, we define its \textit{semantic context} as the pixels that belong to the same semantic human part category of $i$. We use such semantic context to enhance our human representation, as it captures the relations within and among parts. Such enhancement would require a human part label map. Here, we compute a semantic similarity map as a surrogate to expedite computation. Specifically, for each pixel $i$ we compute a semantic similarity map $A^{i} \!\in\![0,1]^{H\!\times\!W}$, where each element $a^{i}_{j} \!\in\!A^i$ stores the `relation' between the \textit{latent} part categories of pixel $i$ and $j$:
\vspace{-0pt}
\begin{equation}\small\label{eq:non_local}
a^{i}_{j} = \frac{1}{z_{i}}exp({\textbf{\textit{h}}_i^{\top}\textbf{\textit{h}}_j}),
\vspace{-0pt}
\end{equation}
where $\textbf{\textit{h}}_i\!\in\!\mathbb{R}^{C}$ and $\textbf{\textit{h}}_j\!\in\!\mathbb{R}^{C}$  are the feature vectors of pixels $i$ and $j$ in $\textbf{\textit{H}}$, respectively. $z_{i}$ is a normalization term: $z_{i}\!=\!\sum_{j}exp({\textbf{\textit{h}}_i^{\top}\textbf{\textit{h}}_j})$. Here $A^i$ can be considered as a soft label map for the semantic human part of $i$.

Then for a pixel $i$, we collect information from its semantic context according to $A^i$:
\vspace{-0pt}
\begin{equation}\small\label{eq:part_enhanced}
\textbf{\textit{c}}_i = \sum\nolimits_{j=1}^{H\times W}\!a^i_j\textbf{\textit{h}}_j \in\mathbb{R}^{C}.
\vspace{-0pt}
\end{equation}
After assembling all the semantic context information for all the parts (pixels) within $\textbf{\textit{H}}$, we get a semantic context enhanced feature $\textbf{\textit{C}}\!\in\!\mathbb{R}^{C\!\times\!H\!\times\!W}$, which is used to compute an improved human representation $\bar{\textbf{\textit{H}}}$:
\vspace{-0pt}
\begin{equation}\small\label{eq:human_feature_refined}
\bar{\textbf{\textit{H}}} = \textbf{\textit{H}} + \textbf{\textit{C}} \in \mathbb{R}^{C\times H\times W}.
\vspace{-0pt}
\end{equation}

\noindent\textbf{2)} \textit{Explicit Facial Region Attending.}
Human face is vital for HOI understanding, as it conveys rich information closely tied to underlying attention and intention~\cite{kleinke1986gaze} of humans. There are many interactions that directly involve human face. For example,
humans use \textit{eyes} to \textit{watch TV}, use \textit{mouth} to \textit{eat food}, and so on. Besides, face-related interactions are typically fine-grained and combined with heavy occlusions on the interacted objects, \eg, \textit{call a phone}, \textit{play a phone}, posing great difficulties for HOI models. To address the above issues, we propose another feature-enhancement mechanism, called explicit facial region attending. This mechanism enriches the object representation $\textbf{\textit{O}}$ via two attention mechanisms:
\vspace*{-0pt}
\begin{itemize}[leftmargin=*]
	\setlength{\itemsep}{0pt}
	\setlength{\parsep}{-2pt}
	\setlength{\parskip}{-0pt}
	\setlength{\leftmargin}{-15pt}
	\vspace{-5pt}
	\item  \textit{Face-aware Attention.} For a human-object pair $(h,o)$, we detect the facial region using an off-the-shelf face detector~\cite{deng2019retinaface}. Then we get an RoIAlign feature $\textbf{\textit{F}}\!\in\!\mathbb{R}^{C\!\times\!H\!\times\!W}$ from the detected facial region as the face representation. An attention score $\alpha\!\in\![0,1]$ is learned for interpreting the importance of the facial region for the object $o$:
\vspace{-3pt}
\begin{equation}\small\label{face-aware}
\alpha = \sigma(\text{FC}_{\times 2}([\textbf{\textit{F}}, \textbf{\textit{O}}])),
\vspace{-3pt}
\end{equation}
where $\sigma$ is the \textit{sigmoid} function, and $\text{FC}_{\times 2}$ stands for two stacked FC layers.
\item \textit{Face-agnostic Attention.} The face-aware enhancement addresses the relevance between human face and object. To mine the potential relations between object and other human regions, we propose a face-agnostic attention. We first remove the facial region from the human $h$, by setting the pixel values in the face region to zero. Then we get the corresponding RoIAlign feature $\bar{\textbf{\textit{F}}}\!\in\!\mathbb{R}^{C\!\times\!H\!\times\!W}$ from the face-removed human regions. Finally, we calculate an importance score $\bar{\alpha}\!\in\![0,1]$ between $\bar{\textbf{\textit{F}}}$ and $\textbf{\textit{O}}$:
\vspace{-7pt}
\begin{equation}\small\label{face-agnostic}
\bar{\alpha} = \sigma(\text{FC}_{\times 2}([\bar{\textbf{\textit{F}}}, \textbf{\textit{O}}])).
\vspace{-7pt}
\end{equation}
\vspace*{-10pt}
\end{itemize}
Considering Eqs.~\!(\ref{face-aware},~\!\ref{face-agnostic}), the object feature ${\textbf{\textit{O}}}$ is enhanced by:
\vspace{-4pt}
\begin{equation}\small\label{face}
\bar{\textbf{\textit{O}}} = \textbf{\textit{O}} + \alpha\textbf{\textit{F}} + \bar{\alpha}\bar{\textbf{\textit{F}}}\in \mathbb{R}^{C\times H\times W}.
\vspace{-0pt}
\end{equation}

In our cascade framework, for a human-object pair $(h,o)\!\in\!\mathcal{O}^t\!\times\!\mathcal{O}^t$ at stage $t$, we update its visual feature $\textbf{\textit{X}}^t_{\text{v}}\!\in\!\mathbb{R}^{3C\times H\times W}$ by considering the one $\textbf{\textit{X}}^{t-1}_{\text{v}}\!\in\!\mathbb{R}^{3C\times H\times W}$
in prior stage:
\vspace{-2pt}
\begin{equation}\small
\bar{\textbf{\textit{X}}}{}^t_{\text{v}} =\text{FC}_{\times 2}(\textbf{\textit{X}}^t_{\text{v}}+\textbf{\textit{X}}^{t-1}_{\text{v}})\in\mathbb{R}^{1024}.
\vspace{-2pt}
\end{equation}
We do not update semantic $\textbf{\textit{X}}_{\text{s}}$ and geometric $\textbf{\textit{X}}_{\text{g}}$ features.

\vspace{-4pt}
\subsubsection{Relation Ranking Module}\label{rrn}
\vspace{-2pt}
Once obtaining the features $\{\textbf{\textit{X}}_{\text{s}}, \textbf{\textit{X}}_{\text{g}}, \bar{\textbf{\textit{X}}}_{\text{v}}\}$ of a human-object pair, we can directly predict its action label. However, a big issue here is how to sample human-object pairs. Given the proposals detected from the localization network, previous HOI methods typically pair all humans and objects, leading to large computational overhead. As a matter of fact, human beings interact with the world following some regularity rather than in a pure chaotic way~\cite{baldassano2017human}. By leveraging such regularity, we propose a human-object relation ranking module (RRM) to select high-quality HOI candidates for further relation recognition. This also helps decrease the difficulty in relation classification and erase the serious class imbalance, as the samples for `non-interaction' class are much more than the ones of any other interaction classes.

RRM is built upon an insight that, although some human-object relations are miss annotated in HOI datasets, the annotated human-object pairs tend to be more relevant (\ie, higher ranking score) than those without any HOI relation labelling.  Given the detection results $\mathcal{O}$ of the instance localization network $\mathtt{L}$ (\S\ref{oln}), we denote the set of all the possible human-object pairs as: $\mathcal{P}\!=\!\{P=(h, o)\!\in\!\mathcal{O}\!\times\!\mathcal{O}\}$. $\mathcal{P}$ can be further divided into two subsets: $\mathcal{P}=\hat{\mathcal{P}}\!\cup\!\check{\mathcal{P}}$, where $\hat{\mathcal{P}}$ and $\check{\mathcal{P}}$ indicate the sets of annotated and un-annotated human-object pairs, respectively.
The goal of RRM is to learn a ranking function $\texttt{g}: \mathbb{R}^{1024+256}\!\rightarrow\!\mathbb{R}$ that fulfills the following constraint:
\vspace{-1pt}
\begin{equation}\small
 \forall \hat{P} \succ \check{P}:	\texttt{g}(\hat{P}) > \texttt{g}(\check{P}),  ~~~~\text{where~~} \hat{P}\!\in\!\hat{\mathcal{P}}, \check{P}\!\in\!\check{\mathcal{P}}.
\vspace{-1pt}
\end{equation}
Here $\hat{P}\!\succ\!\check{P}$ means $\hat{P}$ has a higher ranking than $\check{P}$ . $\texttt{g}(P)$ gives the ranking score of $P$:
\vspace{-1pt}
\begin{equation}\small
\texttt{g}(P)= \sigma(\text{FC}_{\times 1}(\bar{\textbf{\textit{X}}}_{\text{v}}, \textbf{\textit{X}}_{\text{g}}))\in[0,1].
\vspace{-1pt}
\end{equation}
In RRM, the learning of $\texttt{g}$ is achieved by minimizing the following pairwise ranking hinge loss:
\vspace{-1pt}
\begin{equation}\small\label{ranking}
\mathcal{L}_{\text{RRM}} = \sum\nolimits_{\hat{P}\in\hat{\mathcal{P}}}\sum\nolimits_{\check{P}\in\check{\mathcal{P}}}\text{max}(0, \texttt{g}(\check{P}) - \texttt{g}(\hat{P}) + \epsilon)),
\vspace{-1pt}
\end{equation}
where the margin $\epsilon$ is empirically set as 0.2. This loss penalizes the situation that assigning an un-annotated pair $\check{P}$ with a higher ranking score, compared to a labeled pair $\hat{P}$.



\vspace{-5pt}
\subsubsection{Relation Classification Module}\label{icn}
\vspace{-2pt}
Through RRM, only a few top-ranked, high-quality human-object pairs are preserved and fed into a triple-stream~\cite{zhang2019graphical}, relation classification module (RCM) for final HOI recognition. For a HOI candidate ($h, o$), the semantic $\textbf{\textit{X}}_{\text{s}}$, geometric $\textbf{\textit{X}}_{\text{g}}$ and visual $\bar{\textbf{\textit{X}}}_{\text{v}}$  features, are separately fed into a corresponding stream in RCM for estimating a HOI action score vector independently:
\vspace{-4pt}
\begin{align}\small\label{vis}
\begin{split}
\text{semantic stream:}& ~~\textbf{\textit{s}}_{\text{s}} = \sigma(\text{FC}_{\times 1}(\textbf{\textit{X}}_{\text{s}})) \in [0,1]^N, \\
\text{geometric stream:}& ~~\textbf{\textit{s}}_{\text{g}} = \sigma(\text{FC}_{\times 1}(\textbf{\textit{X}}_{\text{g}})) \in [0,1]^N, \\
\text{visual stream:} & ~~\textbf{\textit{s}}_{\text{v}} = \sigma(\text{FC}_{\times 1}(\bar{\textbf{\textit{X}}}_{\text{v}})) \in [0,1]^N,
\end{split}
\vspace{-10pt}
\end{align}
where $\textbf{\textit{s}}_{\text{s}}$, $\textbf{\textit{s}}_{\text{g}}$ and $\textbf{\textit{s}}_{\text{v}}$ are the score vectors from semantic, geometric and visual streams, respectively, and $N$ is the number of pre-defined actions in HOI. Note that here follows a multi-label classification setting.

During training, for each stream, the binary cross-entropy loss is used to evaluate the discrepancy between the output score and truth target. The total loss  $\mathcal{L}_{\text{RCM}}$ is the sum of the ones from streams.
During inference, the final prediction is obtained by:
\vspace{-3pt}
\begin{equation}\small\label{geo}
\textbf{\textit{s}} = (\textbf{\textit{s}}_{\text{v}}+\textbf{\textit{s}}_{\text{g}}) \odot \textbf{\textit{s}}_{\text{s}},
\vspace{-2pt}
\end{equation}
where $\odot$ denotes the Hadamard product.

\subsection{Relation Segmentation}\label{relation_segmentation}
So far, we strictly follow the classic \textit{relation detection} setting in HOI recognition~\cite{gkioxari2018detecting,li2019transferable,fang2018pairwise,xu2019learning}, \ie, identify the interaction entities by bounding boxes.  Now we focus on  how to adapt our cascade framework to \textit{relation segmentation}, which addresses more fine-grained HOI understanding by representing each entity at the pixel level.

Inspired by~\cite{cai2018cascade}, for the instance localization network $\mathtt{L}^t$ at each stage $t$, an instance segmentation head $\mathtt{S}^t$ is added and the whole workflow (Eqs.~(\ref{eq:y}, \ref{eq:o})) is changed as:
\vspace{-3pt}
\begin{equation}\small
\begin{aligned}
	\!\!\!\!\text{Instance Detection:} ~~\textbf{\textit{Y}}^t &= \mathtt{P}(\textbf{\textit{I}}, o^{t-1}), o^t~= \mathtt{D}^t(\textbf{\textit{Y}}^t),\\
    \!\!\!\!\text{Instance Segmentation:} ~~\bar{\textbf{\textit{Y}}}^t &= \mathtt{P}(\textbf{\textit{I}}, o^t), ~~~~\bar{o}^t~= \mathtt{S}^t(\bar{\textbf{\textit{Y}}}^t, \bar{\textbf{\textit{Y}}}^{t-1}),\!\!\!\!
	\end{aligned}
	\vspace{-3pt}
\label{eq:i}
\end{equation}
where$_{\!}$ $\bar{o}^{t\!}\!\in\!\bar{\mathcal{O}}^{t\!}$ indicates$_{\!}$ a$_{\!}$ generated$_{\!}$ object$_{\!}$ instance$_{\!}$ mask.$_{\!}$ Then, in our relation recognition network (\S\ref{irn}), the human-object pair $(h,o)$ is sampled from the object masks $\bar{\mathcal{O}}^t$ and associated with finer features: $\textbf{\textit{H}}$, $\textbf{\textit{O}}$ and $\textbf{\textit{U}}$ by pixel-wise RoI. In addition, the generation of geometric feature $\textbf{\textit{X}}_{\text{g}}$ is based on pixel-level masks. The binary cross-entropy loss $\mathcal{L}_{\text{SEG}}^t$ is used for training $\mathtt{S}$.

\begin{figure*}
	\begin{minipage}[t]{0.35\textwidth}
		\centering
		\begin{threeparttable}
			\resizebox{1\textwidth}{!}{
				\setlength\tabcolsep{15pt}
				\renewcommand\arraystretch{1.065}
				\begin{tabular}{c|c||c}
					\hline\thickhline
					\rowcolor{mygray}
					Challenge & Team & $\text{mAP}_{rel}$ \\ \hline \hline
					& Ours   &  \textbf{66.04} \\
					 & GMVM   &  60.26  		\\
					PIC$_{19}$ Challenge	& FINet  &  56.93  		\\
					(\textit{HOIW Track})& F2INet &  49.13 			\\
					& TIN~\cite{li2019transferable}   &  48.64 			\\
                    \hline
				\end{tabular}
			}
		\end{threeparttable}
		\vspace*{-8pt}
		\captionsetup{font=small}
		\makeatletter\def\@captype{table}\makeatother\caption{\small\textbf{Relation detection results on HOIW \texttt{test} set in PIC$_{19}$ Challenge} (\S\ref{sec:41}). 
			\label{table:hoiw}}
		\vspace*{-18pt}
	\end{minipage}
	\begin{minipage}[t]{0.65\textwidth}
		\centering
		\begin{threeparttable}
			\resizebox{1\textwidth}{!}{
				\setlength\tabcolsep{10pt}
				\renewcommand\arraystretch{1.02}
				\begin{tabular}{c|c||ccc|c}
					\hline\thickhline
					\rowcolor{mygray}
					& & R@100  & R@100& R@100 &  \\
					\rowcolor{mygray}
					\multirow{-2}{*}{Challenge} & \multirow{-2}{*}{Team}& mIoU: 0.25&mIoU: 0.50&mIoU: 0.75 &\multirow{-2}{*}{Mean} \\ \hline \hline
					& Ours & \textbf{60.17} & \textbf{55.11} & \textbf{42.29} & \textbf{52.52} \\
					PIC$_{19}$ Challenge  & HTC+iCAN & 56.21 		& 52.32 	& 37.49		& 48.67 \\
					(\textit{PIC Track})  & RelNet 		& 53.17  	 & 49.26 	& 32.44	 	& 44.96 \\
					 & XNet 			 & 38.42	 & 33.15  	& 17.29 	& 29.62\\
                    \hline
				\end{tabular}
			}
			\vspace*{-8pt}
			\captionsetup{font=small}
			\makeatletter\def\@captype{table}\makeatother\caption{\small\textbf{Relation segmentation results on PIC \texttt{test} set in PIC$_{19}$ Challenge.}  Please see \S\ref{sec:41} for details.
				\label{table:pic}}
		\end{threeparttable}
	\end{minipage}
	\vspace*{-8pt}
\end{figure*}

\begin{figure*}[t]
	\begin{center}
		\includegraphics[width=\linewidth]{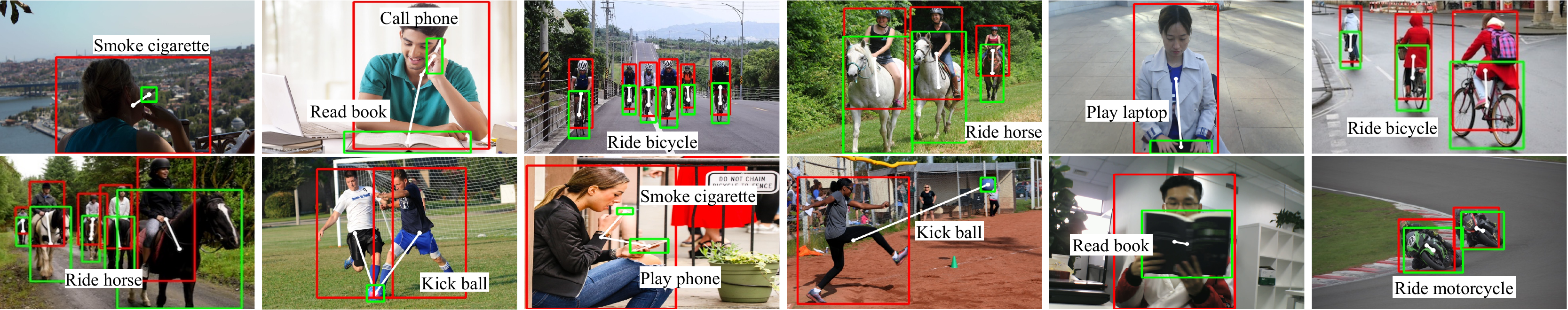}
	\end{center}
	\vspace{-16pt}
	\captionsetup{font=small}
	\caption{\small\textbf{Visual results for relation detection, on HOIW \texttt{test} set in PIC$_{19}$ Challenge} (\S\ref{sec:41}).
	}
	\label{fig:hoiw}
	\vspace{-13pt}
\end{figure*}

\subsection{Implementation Details}
\noindent\textbf{Training Loss.}
Since all the modules mentioned above are differentiable, our cascade architecture can be trained in an end-to-end manner.
In the \textit{relation detection} setting, the entire loss is computed as:
\vspace{-1pt}
\begin{equation}\small\label{eq:loss}
\mathcal{L} = \sum\nolimits_{t=1}^T \beta^t\mathcal{L}_{\text{LOC}}^t + \gamma^t(\mathcal{L}_{\text{RRM}}^t + \mathcal{L}_{\text{RCM}}^t).
\vspace{-1pt}
\end{equation}
Here, $\mathcal{L}^t_{\text{LOC}}$ is the localization loss at stage $t$ (\S\ref{irn}). $\mathcal{L}_{\text{RRM}}^t$ and $\mathcal{L}_{\text{RCM}}^t$ are the losses of RRM (\S\ref{rrn}) and RCM (\S\ref{icn}), respectively. The coefficients $\beta_t$ and $\gamma_t$ are used to balance the contributions of different stages and tasks. There are three stages used in our method ($T\!=\!3$), and we set $\beta\!=\!\gamma\!=\![1, 0.5, 0.25]$.
In the \textit{relation segmentation} setting, the instance segmentation head $\mathtt{S}^t$ is injected into the network (\S\ref{relation_segmentation}). The corresponding instance segmentation loss $\mathcal{L}_{\text{SEG}}^t$ is further added in Eq.~(\ref{eq:loss}), with coefficients $[1, 0.5, 0.25]$.

\noindent\textbf{Cascade Inference.} During inference, the object proposals generated by the instance localization network in different stages are merged together. We remove the ones whose confidence scores are smaller than 0.3. Then, all the possible human-object pairs, generated from the remaining proposals, are fed into RRM for relation ranking. After that, we only select the top 64 pairs as candidates and feed them into RCM for final relation classification. The last-stage output of RCM is used as the final action score.

\vspace{-2pt}
\section{Experiments}
\vspace{-1pt}
Experiments are conducted on three datasets, \ie, HOIW, PIC and V-COCO~\cite{gupta2015visual}. The former two are from the PIC$_{19}$ Challenge, and the last one is a gold standard benchmark. 

\noindent\textbf{Training Settings:} Unless specially noted, we adopt the following training settings for all the experiments.
We use ResNet-50~\!\cite{he2016deep} as the backbone. The training includes two phases: 1) training the instance localization network; and then 2) jointly training the instance localization and interaction recognition networks. In the first phase, the network is initialized using the weights pre-trained on COCO~\!\cite{lin2014microsoft}. The three stages are trained using gradually increased IoU thresholds $\mu_{\!}\!=_{\!}\!\{0.5,0.6,0.7\}$~\!\cite{cai2018cascade,chen2019hybrid}.
Training images are resized to a maximum scale of $1333\!\times\!800$, without changing the aspect ratio.
We apply horizontal flipping for data augmentation and train the network for 12 epochs with batch size 16 and initial learning rate 0.02, which is reduced by 10 at epoch 8 and 11.
In the second phase, we adopt the image-centric training strategy~\cite{girshick2015fast}, \ie, using pairwise samples from one image to make up a mini-batch. For each mini-batch, we sample at most 128 HOI proposals with a ratio of 1:3 of positive to negative samples to jointly train RRM and RCM. At each stage, the same IoU threshold $\mu$ is used to determine positive HOI proposals so that the training data for the interaction recognition network closely match the detection quality. Besides, ground-truth HOIs are also used at each stage for training. The second phase is trained with learning rate 0.02 and batch-size 8 for 7 epochs.

\noindent\textbf{Reproducibility:} Our model is implemented on PyTorch and trained on 8 NVIDIA Tesla V100 GPUs with a 32GB memory per-card. Testing is conducted on a single NVIDIA TITAN Xp GPU with 12 GB memory.

\begin{figure*}[t]
	\begin{center}
		\includegraphics[width=\linewidth]{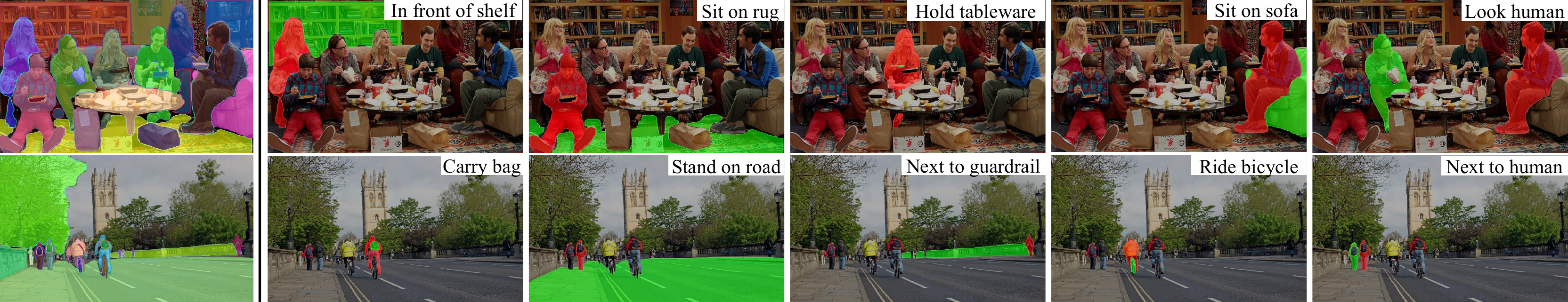}
	\end{center}
	\vspace{-18pt}
	\captionsetup{font=small}
	\caption{\small\textbf{Visual results for relation segmentation, on PIC \texttt{test} set in PIC$_{19}$ Challenge} (\S\ref{sec:41}). First column: Instance segmentation results. Last five columns: Top ranked $_{\!}\left\langle\textit{{human}, {verb}, {object}}\right\rangle_{\!}$ triplets. For each triplet, the \textit{human} and \textit{object} are shown in {\color{red}red} and {\color{green}green}. 
	}
	\label{fig:pic}
	\vspace{-8pt}
\end{figure*}

\vspace{-0pt}
\subsection{Results on PIC$_{19}$ Challenge}
\label{sec:41}
\vspace{-1pt}
\noindent\textbf{Dataset:} The PIC$_{19}$ Challenge includes two tracks, \ie, HOIW and PIC tracks, each with a standalone dataset:
\begin{itemize}[leftmargin=*]
	\setlength{\itemsep}{0pt}
	\setlength{\parsep}{-2pt}
	\setlength{\parskip}{-0pt}
	\setlength{\leftmargin}{-15pt}
	\vspace{-3pt}
	\item \textbf{HOIW}~\cite{liao2019ppdm} is for human-object relation detection. It has 29,842 training and 8,794 testing images, with bounding box annotations for 11 object and 10 action categories. Since it does not provide \texttt{train}/\texttt{val} splits, in our ablation study, we randomly choose 9,999 images for \texttt{val} and the other 19,843 for \texttt{train}; for the challenge result, we use \texttt{train+val} for training.
	\item \textbf{PIC} is for human-object relation segmentation. It has 17,606 images (12,654 for \texttt{train}, 1,977 for \texttt{val} and 2,975 for \texttt{test}) with pixel-level annotations for 143 objects. It covers 30 relationships, including 6 geometric (\eg, \textit{next-to}) and 24 non-geometric (\eg, \textit{look}, \textit{talk}).
\vspace*{-3pt}
\end{itemize}
\noindent\textbf{Evaluation Metrics:}
Standard evaluation metrics in the challenges are adopted.
For HOIW, the performance is evaluated by $\text{mAP}_{rel}$. A detected triplet $_{\!}\left\langle\textit{{human}, {verb}, {object}}\right\rangle_{\!}$ is considered as a true positive if the predicted \textit{verb} is correct and both the \textit{human} and \textit{object} boxes have IoUs at least 0.5 with the corresponding ground-truths.
For PIC, we use Recall@100 (R@100), which is averaged over two relationship categories (\ie, \textit{geometric} and \textit{non-geometric}) and three IoU thresholds (\ie, 0.25, 0.5 and 0.75). In our ablation study, we also consider R@50 and R@20  to measure the performance under stricter conditions.

\begin{figure*}[t]
	\begin{center}
		\includegraphics[width=\linewidth]{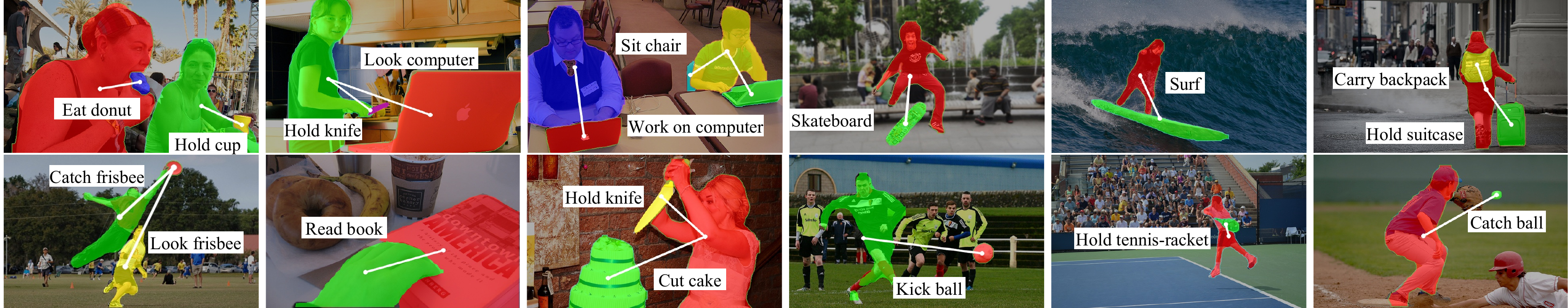}
	\end{center}
	\vspace{-15pt}
	\captionsetup{font=small}
	\caption{\small\textbf{Visual results for relation segmentation, on V-COCO \texttt{test} set~\cite{gupta2015visual}.} See \S\ref{sec:42} for details.
}
	\label{fig:vcoco}
	\vspace{-15pt}
\end{figure*}

\noindent\textbf{Performance on the HOIW Track:} Our approach reaches the $1^{st}$ place for relation detection on the HOIW track. As reported in \tabref{table:hoiw}, our result is substantially better than other teams. In particular, it is \textbf{5.78\%} absolutely better than the $2^{nd}$ (GMVM) and \textbf{9.11\%} better than the $3^{rd}$ (FINet). Our approach also significantly outperforms one published state-of-the-art, \ie, TIN~\cite{li2019transferable} .
\figref{fig:hoiw} presents some visual results on HOIW \texttt{test}. Our model shows robust to various challenges, \eg, occlusions, subtle relationships, \etc. 


\noindent\textbf{Performance on the PIC Track:} Our approach also reaches the $1^{st}$ place for relation segmentation on the PIC track.
As reported in \tabref{table:pic}, our overall score (\textbf{52.52\%}) outperforms the $2^{nd}$ place by \textbf{3.85\%} and the $3^{rd}$ by \textbf{7.56\%}. 
\figref{fig:pic} depicts visual results of two complex scenes on PIC \texttt{test}. Our method shows outstanding performance in terms of instance segmentation as well as interaction recognition. It can identify both geometric and non-geometric relationships, and is capable of recognizing many fine-grained interactions, \eg, \textit{look human}, \textit{hold tableware}. In this track, the instance localization network is instantiate as Eq.~\!(\ref{eq:i}). 

\vspace{-0pt}
\subsection{Results on V-COCO}
\label{sec:42}
\vspace{-1pt}
\noindent\textbf{Dataset:} V-COCO~\cite{gupta2015visual} provides verb annotations for MS-COCO~\cite{lin2014microsoft}. Proposed in 2015, it is the first large-scale dataset for HOI understanding and remains the most popular one today. It contains 10,346 images in total ($2,533/2,867/4,946$ for \texttt{train}/\texttt{val}/\texttt{test} splits). 16,199 human instances are annotated with 26 action labels, wherein three actions (\ie, \textit{cut, hit, eat}) are annotated with two types of targets (\ie, instrument and direct object).

\noindent\textbf{Evaluation Metrics:} We use the original role mean AP ($\text{mAP}_{role}$), which is exactly same with $\text{mAP}_{rel}$ in HOIW.

\noindent\textbf{Performance:}
Since V-COCO has both bounding box and mask annotations, we provide two variants of our methods, \ie, $Ours_{\text{bbox}}$ and $Ours_{\text{mask}}$, where $Ours_{\text{bbox}}$ is trained with box annotations while $Ours_{\text{mask}}$ uses groundtruth masks. For fairness, during evaluation, the mask outputs of $Ours_{\text{mask}}$ are transformed to boxes.
\tabref{table:vcoco} summarizes the results in comparison with  8 state-of-the-arts. $Ours_{\text{bbox}}$ outperforms TIN~\cite{li2019transferable} by $\textbf{0.5\%}$ and RPNN~\cite{Zhou_2019_ICCV} by $\textbf{0.8\%}$. $Ours_{\text{mask}}$ further improves $Ours_{\text{bbox}}$ by $\textbf{0.6\%}$, which suggests the superiority of the mask-level representation over the box-level. We would like to note that \cite{wan2019pose} reported a $52.0\%$ $\text{mAP}_{role}$ on V-COCO. However, It relies on an expensive pose estimator, thus it is unfair to directly compare with our method. Without the pose estimator,
\cite{wan2019pose} obtains a score of $48.6\%$, slightly worse than $Ours_{\text{mask}}$.
\begin{table}
	\centering
	\small
	\resizebox{0.49\textwidth}{!}{
	\setlength\tabcolsep{8pt}
	\renewcommand\arraystretch{1.0}
	\begin{tabular}{r|c|l||c}
		\hline\thickhline
		\rowcolor{mygray}
		 Methods &\!\!Publication\!\!& ~~Backbone & $\text{mAP}_{role}(\%)$ \\ \hline \hline
		Gupta \textit{et. al.}~\cite{gupta2015visual} & Arxiv15 & ResNet-50-FPN &  31.8  	\\
		Interact~\cite{gkioxari2018detecting}      & CVPR18  & ResNet-50-FPN &  40.0		\\
		GPNN~\cite{qi2018learning}					  & ECCV18  & ResNet-50     &  44.0		\\
		iCAN~\cite{gao2018ican} 		              & BMVC18  & ResNet-50     &  45.3 	\\
		Xu \textit{et. al.}~\cite{xu2019learning}     & CVPR19  & ResNet-50-FPN &  45.9 	\\
		Wang \textit{et. al.}~\cite{wang2019deep}     & ICCV19  & ResNet-50     &  47.3 	\\
		RPNN~\cite{Zhou_2019_ICCV} 	                  & ICCV19  & ResNet-50     &  47.5 	\\
		TIN~\cite{li2019transferable} & CVPR19  & ResNet-50     &  47.8 	\\ \hline
		\!\!$Ours_{\text{bbox}}$							  & -	    & ResNet-50	    &  48.3		\\
		\!\!$Ours_{\text{mask}}$						  & -	    & ResNet-50	    &  \textbf{48.9}		\\ \hline
		\end{tabular}
	}
	\captionsetup{font=small}
	\caption{\small\textbf{Comparison of $\text{mAP}_{role}$ on V-COCO \texttt{test}~\cite{gupta2015visual}} (\S\ref{sec:42}). }
	\vspace{-15pt}
	\label{table:vcoco}
\end{table}
In \figref{fig:vcoco}, we illustrate HOI segmentation results of $Our_{\text{mask}}$ on V-COCO \texttt{test} set. It precisely recognizes many fine-grained interactions, such as \textit{look computer}, \textit{read book}, etc.

Overall, our model consistently achieves promising results over different datasets as well as two different settings (\ie, relation detection and segmentation), which clearly reveals its remarkable performance and strong generalization.

\begin{figure}
	\begin{center}
		\includegraphics[width=\linewidth]{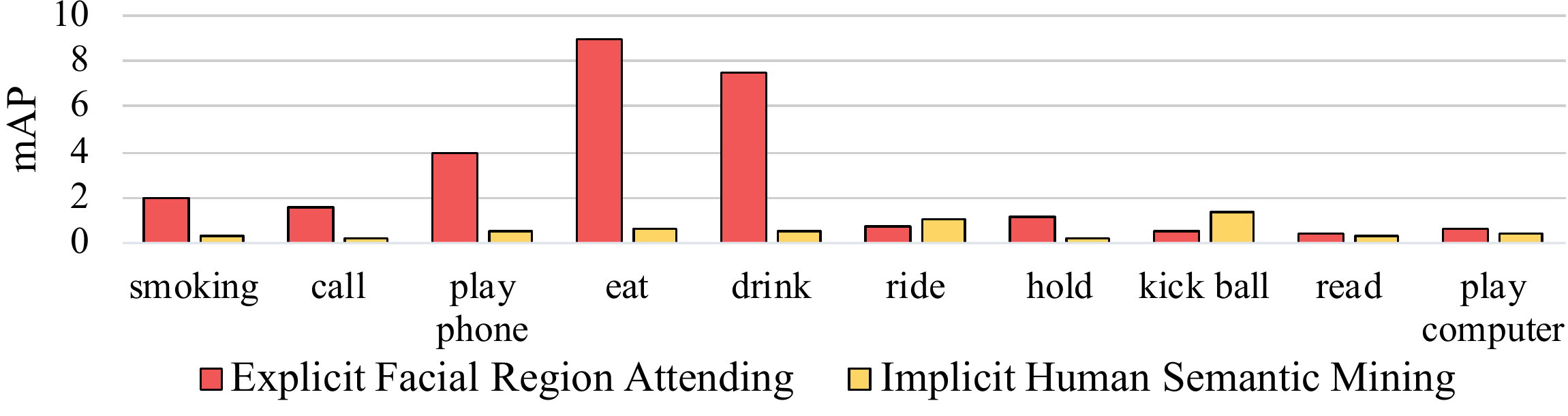}
	\end{center}
	\vspace{-15pt}
	\captionsetup{font=small}
	\caption{\small\textbf{Per-category performance improvement $\Delta \text{mAP}_{rel}$ of the proposed attention modules on HOIW \texttt{val} set} (\S\ref{sec:43}). }
	\vspace{-8pt}
	\label{fig:fa-lpa}
\end{figure}

\begin{table}[t]
	\centering
	\small
	\resizebox{0.49\textwidth}{!}{
		\setlength\tabcolsep{6pt}
		\renewcommand\arraystretch{1.0}
		\begin{tabular}{cccc||ccc|c}
			\hline\thickhline
			\rowcolor{mygray}
			& & & &  \multicolumn{3}{c|}{PIC} & HOIW \\ \cline{5-8}
			\rowcolor{mygray}
			\multirow{-2}{*}{IHSM} &\multirow{-2}{*}{EFRA} & \multirow{-2}{*}{RRM} & \multirow{-2}{*}{CAS} &R@20 & R@50 & R@100 & $\text{mAP}_{rel}$ \\ \hline \hline
			\ding{55} & \ding{55} & \ding{55} & \ding{55} & 17.0 & 28.0 & 33.9 & 33.9\\    \hline
			\ding{51} & & & 				 & 17.9 & 28.6 & 34.3 & 34.4 \\
			& \ding{51}& & 					 & 17.6 & 27.5 & 34.6&  36.7\\
			\ding{51}& \ding{51}& & 		 & 18.5 & 28.3 & 35.4 & 37.5\\
			\ding{51}& \ding{51}& \ding{51}& & 19.0 & 28.9& 35.9 & 38.6\\
			\ding{51}& \ding{51}& \ding{51}& \ding{51}& \textbf{27.8} & \textbf{38.3} & \textbf{45.3} & \textbf{43.7} \\  \hline
		\end{tabular}
	}
	\captionsetup{font=small}
	\caption{\small\textbf{\!Ablation study of key components} in our cascade model. }
	\label{table:component}
	\vspace{-12pt}
\end{table}

\vspace{-3pt}
\subsection{Ablation Study}\label{sec:ablation_study}
\label{sec:43}
\vspace{-2pt}
\noindent\textbf{Key Component Analysis.}
First, we investigate the influence of essential components in our framework, \ie, implicit human semantic mining (IHSM), explicit facial region attending (EFRA), relation ranking module (RRM) and cascade network architecture (CAS). We first build a baseline model without any of these components, and then gradually add each into the baseline for investigation. As reported in \tabref{table:component}, all these components can improve the performance in both PIC and HOIW datasets. \textbf{1)} IHSM and EFRA help to learn more discriminative visual features and further boost the performance (\eg, \textbf{0.5\%} and \textbf{2.8\%} performance improvements on HOIW). \textbf{2)} \figref{fig:fa-lpa} shows the per-category performance improvement of IHSM and EFRA on HOIW \texttt{val} set. Obviously, EFRA improves the performance on face-related interactions (\eg, \textit{eat, drink, smoking, call}) and discriminates these categories from some similar ones, \eg, \textit{play phone}. In contrast, IHSM is more effective for the actions with specific poses, \eg, \textit{ride, kick ball}. \textbf{3)} RRM plays a key role in pruning negative human-object pairs, as proved by \tabref{table:component}. Moreover, RRM improves the average inference speed by about \textbf{$80$ms} on HOIW. \textbf{4)} Our cascade architecture substantially boosts the performance, \ie, \textbf{8.8\%} absolute improvement in PIC and \textbf{5.1\%} in HOIW.

\noindent\textbf{Cascade Architecture Analysis.}
We study the impact of the number of stages $T$ used in our cascade network by varying it from $1$ to $5$.
The IoU thresholds used for these five stages are $[0.5, 0.6, 0.7, 0.75, 0.8]$. The results in
\tabref{table:cascade_num} show that the performance is significantly improved by adding a second stage, \ie, \textbf{6.5\%} in terms of R@20 in PIC and \textbf{3.5\%} in terms of $\text{mAP}_{rel}$ in HOIW. When further adding more than 3 stages, the performance gain is marginal. \tabref{table:cascade_num} also reports the average inference time for these variants on HOIW \texttt{val} set. The test speed decreases with adding more stages and drops quickly after using $4$ or $5$ stages. Considering the model complexity and performance, we choose $T\!=\!3$ as our default setting.
\tabref{table:cascade} reports the performance comparison of our approach with ($T\!=\!3$) or without ($T\!=\!1$) cascade under different backbones, \ie, ResNet-50, ResNet-101 and ResNeXt-101.
The results reveal that our cascade network consistently improves the performance on various backbones.

\begin{table}[t]
	\centering
	\small
	\resizebox{0.47\textwidth}{!}{
		\setlength\tabcolsep{8pt}
		\renewcommand\arraystretch{1.0}
		\begin{tabular}{c||c||ccc|c}
			\hline\thickhline
			\rowcolor{mygray}
			&  & \multicolumn{3}{c|}{PIC} &  HOIW \\ \cline{3-6}
			\rowcolor{mygray}
			\multirow{-2}{*}{$T$}	& \multirow{-2}{*}{Speed (ms)} & R@20 & R@50 & R@100 &  $\text{mAP}_{rel}$ \\  \hline \hline
			1 & 145 & 19.0 & 28.9 & 35.9 & 38.6 \\
			2 & 163 & 25.5 & 36.4 & 43.8 & 42.1 \\
			3 & 198 & \textbf{27.8} & \textbf{38.3} & \textbf{45.3} & \textbf{43.7} \\
			4 & 253 & \textbf{27.8} & \textbf{38.3} & 45.2 & \textbf{43.7}  \\
			5 & 314 & 27.6  & 38.1 &  45.2 & 43.4\\ \hline
		\end{tabular}
	}	
	\captionsetup{font=small}
	\caption{\small\textbf{\!Impact of the number of stages $T$} in our cascade model.}
	\label{table:cascade_num}
	\vspace{-9pt}
\end{table}

\begin{table}[t]
	\centering
	\small
	\resizebox{0.49\textwidth}{!}{
		\setlength\tabcolsep{6pt}
		\renewcommand\arraystretch{1.0}
		\begin{tabular}{l|c||ccc|c}
			\hline\thickhline
			\rowcolor{mygray}
			&  	 & \multicolumn{3}{c|}{PIC} &  HOIW \\ \cline{3-6}
			\rowcolor{mygray}
			\multirow{-2}{*}{Backbone}	& \multirow{-2}{*}{Cascade}& R@20 & R@50 & R@100 &  $\text{mAP}_{rel}$ \\  \hline \hline
			ResNet-50 		& \ding{55} & 19.0 & 28.9 & 35.9 & 38.6\\
			ResNet-50 		& \ding{51} & 27.8 & 38.3 & 45.3 & 43.7\\ \hline
			ResNet-101 		& \ding{55} & 20.8 & 31.4 & 38.9 & 40.2\\
			ResNet-101 		& \ding{51} & 28.6 & 39.8 & 47.0 & 44.4\\ \hline
			ResNeXt-101 	& \ding{55} & 22.9 & 34.3 & 42.6 & 44.2\\
			ResNeXt-101 	& \ding{51} & 29.6 & 41.2 & 48.9 & 48.2\\ \hline
		\end{tabular}
	}	
	\captionsetup{font=small}
	\caption{\small\textbf{Ablation study of the cascade architecture with various backbones.}}
	\label{table:cascade}
	\vspace{-15pt}
\end{table}

\noindent\textbf{Efficacy of Our Relation Representation and Score Fusion Strategy.}
In our method, three kinds of features, $\textbf{\textit{X}}_{\text{s}}$,  $\textbf{\textit{X}}_{\text{g}}$ and $\textbf{\textit{X}}_{\text{v}}$, are used for capture semantic, geometric and visual information for relation modeling.
\tabref{table:rcm} reports the performance with only considering one single feature. As seen, the visual feature is more important than the other two. In addition, we further investigate different ways to fuse the action scores from the three features, we find that the one used in Eq.~\eqref{geo} is the best.

\noindent\textbf{Exploring Better Relation Representation.}
Existing HOI methods typically use coarse bounding boxes to represent the entities, however, is it the best choice? To answer this, we perform experiments to explore more powerful relation representation. We evaluate the performance of our model on PIC \texttt{val} set using four different representations: a) BBox; b) Mask; c) BBox+Mask (max); and d) BBox+Mask (sum). Here, a) and b) means that we extract the features $\textbf{\textit{H}}$, $\textbf{\textit{O}}$, $\textbf{\textit{U}}$ by applying RoIAlign over bbox and mask regions, respectively. c) and d) are the fusion of bbox and mask features with element-wise \texttt{max} and \texttt{sum} operations, respectively. Note that the detected entities are the same for all the baselines. The results in \tabref{table:mask} show that mask is superior to bbox, especially under the strictest metric R@20. The two hybrid representations are better than solely using bbox, but slightly worse than the purely mask-based.  In summary,  mask-based representation indeed benefits HOI recognition as it provides more precise information.

\begin{table}[t]
	\centering
	\small
	\resizebox{0.49\textwidth}{!}{
		\setlength\tabcolsep{2pt}
		\renewcommand\arraystretch{1.1}
		\begin{tabular}{c|c||ccc|c}
			\hline\thickhline
			\rowcolor{mygray}
			& & \multicolumn{3}{c|}{PIC} &  HOIW \\ \cline{3-6}
			\rowcolor{mygray}
			\multirow{-2}{*}{Aspect}& \multirow{-2}{*}{Variant}	 & R@20 & R@50 & R@100 &  $\text{mAP}_{rel}$ \\ \hline \hline
			\multirow{3}{*}{\tabincell{c}{Relation\\Representation}}& Semantic Feature ($\textbf{\textit{s}}_{\text{s}}$) 	& 14.5 & 20.0 & 23.3 & 26.5 \\
				& Geometric Feature ($\textbf{\textit{s}}_{\text{g}}$) 	& 19.6 & 26.2 & 32.1 & 30.3 \\
			& Visual Feature ($\textbf{\textit{s}}_{\text{v}}$) 	& 22.2 & 32.8 & 38.2 & 38.1 \\ \hline
			\multirow{3}{*}{\tabincell{c}{Score\\Fusion}} & $\textbf{\textit{s}}_{\text{v}}+\textbf{\textit{s}}_{\text{g}} + \textbf{\textit{s}}_{\text{s}}$ 	& 26.7 & 37.0 & 43.1 & 41.3 \\
			& $\textbf{\textit{s}}_{\text{v}}\odot\textbf{\textit{s}}_{\text{g}} \odot \textbf{\textit{s}}_{\text{s}}$ 				& 27.0 & 37.7 & 43.5 & 41.9 \\
			& $(\textbf{\textit{s}}_{\text{v}}+\textbf{\textit{s}}_{\text{g}}) \odot \textbf{\textit{s}}_{\text{s}}$ 				& \textbf{27.8} & \textbf{38.3} & \textbf{45.3} & \textbf{43.7} \\ \hline
		\end{tabular}
	}
	\captionsetup{font=small}
	\caption{\small\textbf{\!Ablation study of our relation representation and score fusion strategy.\!\!\!}}
	\label{table:rcm}
	\vspace{-10pt}
\end{table}

\begin{table}[t]
	\centering
	\small
	\resizebox{0.49\textwidth}{!}{
		\setlength\tabcolsep{8pt}
		\renewcommand\arraystretch{1.1}
		\begin{tabular}{c||ccc|c}
			\hline\thickhline
			\rowcolor{mygray}
			Relation Representation& R@20 & R@50 & R@100  & Mean\\ \hline \hline
			BBox & 27.1 & 37.9 & 44.8 & 36.6\\
			Mask & \textbf{27.8} & \textbf{38.3} & \textbf{45.3} & \textbf{37.1}\\
			BBox + Mask (\texttt{max}) &27.6 &\textbf{38.3} & 45.1 & 37.0\\
			BBox + Mask (\texttt{sum}) & 27.7&\textbf{38.3} & 45.1 & 37.0\\ \hline
		\end{tabular}
	}	
	\captionsetup{font=small}
	\caption{\small\textbf{Comparison between mask and bbox representations.}}
	\label{table:mask}
	\vspace{-14pt}
\end{table}

\vspace{-2pt}
\section{Conclusion}
\vspace{-1pt}
This paper introduces a cascade network architecture for coarse-to-fine HOI recognition. It consists of an instance localization network and an interaction recognition network, which are densely connected at each stage to fully exploit the superiority of multi-tasking. The interaction recognition network leverages human-centric features to learn better semantics of actions, and comprises two crucial modules for relation ranking and classification. Our model achieves the $1^{st}$ place on both relation detection and relation segmentation tasks in PIC$_{19}$ Challenge, and also outperforms prior methods on a gold standard benchmark, V-COCO. Besides, we empirically demonstrate the advantages of mask over bounding box for more precise relation representation, and will go deep into this in our future research.


{\small
\bibliographystyle{ieee_fullname}
\bibliography{egbib}
}

\end{document}